\documentclass{article}

\usepackage{PRIMEarxiv}

\usepackage[utf8]{inputenc} 
\usepackage[T1]{fontenc}    
\usepackage{hyperref}       
\usepackage{url}            
\usepackage{booktabs}       
\usepackage{amsfonts}       
\usepackage{nicefrac}       
\usepackage{microtype}      
\usepackage{lipsum}
\usepackage{graphicx}
\graphicspath{{media/}}     
\usepackage{listings}
\usepackage{xcolor}
\usepackage{float}
\usepackage{amsthm}
\newtheorem{definition}{Definition}
\usepackage{amsmath}
\usepackage{pifont}

\title{ContinuityBench: A Benchmark and Systems Study of Stateful Failover in Multi-Provider LLM Routing}

\author{
  Vishal Pandey\\
  Metriqual\\
  London, UK\\
  \texttt{vishal@metriqual.com} \\
   \And
  Gopal Singh\\
  Metriqual \\
  Athens, GR\\
  \texttt{gopal@metriqual.com} \\
}

\begin{document}
\maketitle

\begin{abstract}
In production large language model (LLM) deployments, high API availability guarantees do not equate to conversational continuity. When a primary provider experiences an outage or strict rate-limiting, naive stateless failover mechanisms successfully maintain uptime but silently discard conversation history, severely disrupting the user experience. To rigorously quantify and resolve this failure mode, we introduce two novel metrics: Continuity Preservation Rate (CPR) and Continuity Latency Overhead (CLO). We propose a stateful, multi-provider proxy architecture utilizing a History-Forwarding strategy to seamlessly reconstruct conversational state across heterogeneous LLM endpoints during failover events. Furthermore, we release \textit{continuity-bench} \footnote{\url{https://github.com/Vishal-sys-code/continuity-bench}}, an open evaluation harness designed to stress-test context preservation under high-concurrency provider failure conditions. Our empirical evaluation ($N=750$ failover events) demonstrates that our stateful proxy achieves a 99.20\% CPR [95\% CI: 98.27\%, 99.63\%], cleanly transferring deep conversational context to fallback providers, compared to a near-0\% preservation rate for standard stateless architectures. Finally, we characterize failover latency distributions, identifying the critical necessity of asynchronous exponential backoff with jitter to prevent cascading retry storms against strict-limit fallback APIs. Our results provide a principled foundation for building robust, state-preserving multi-model inference systems.
\end{abstract}


\section{Introduction}
\label{sec:introduction}
A voice agent does not crash during a provider failover. It forgets.
Consider the following failure scenario, observed during a routine debugging session that ultimately motivated this work. A primary LLM provider experiences a transient outage lasting eleven seconds. The retry logic performs exactly as designed: the request is rerouted to a secondary provider, which returns a well-formed response. Every health check reports green. The HTTP status code reads 200. By every operational metric in common use, the system has behaved correctly. Yet the user is forced to repeat themselves from the beginning, because the failover model received none of the preceding conversational context. The dashboard says the system is healthy. The user knows otherwise.
This failure mode is pervasive, consequential, and almost entirely unmeasured. The machine learning community has developed rigorous benchmarks for accuracy~\cite{mmlu}, latency~\cite{anyscale-benchmark}, and hallucination rate~\cite{halueval}. No comparable methodology exists for evaluating whether a multi-provider system preserves conversational continuity when a failover event occurs. The implicit industry assumption that a successful HTTP response entails a successful conversation conflates two fundamentally distinct properties: \textit{availability} (the system returned an answer) and \textit{continuity} (the system returned an answer informed by the full preceding context). In interactive voice applications, the loss of continuity manifests as an abrupt conversational reset, functionally equivalent to a hang-up. In agentic pipelines executing multi-step tasks, the consequences are more insidious: the agent continues operating on a truncated context window, producing plausible but incorrect outputs whose root cause the silent loss of state at the failover boundary may not surface until several reasoning steps later, well after the actual damage has been done.
\subsection{The Gap in Existing Infrastructure}
A growing ecosystem of multi-provider LLM gateways has emerged to address the operational challenges of deploying across heterogeneous model endpoints. Systems such as LiteLLM~\cite{litellm}, Portkey~\cite{portkey}, and OpenRouter~\cite{openrouter} provide unified API interfaces, automatic failover routing, load balancing, and cost optimization across providers including OpenAI, Anthropic, Google, and others. These systems have made substantial contributions to operational reliability, and we build upon several of their architectural patterns in this work.
However, existing gateways operate at the \textit{request} level: each API call is treated as an independent, stateless transaction. When a failover is triggered, the gateway routes the current request to an alternative provider, but the conversational history that preceded it maintained only by the now-unavailable primary is not reconstructed. The failover model receives a single decontextualized message rather than the full dialogue. This is not an implementation oversight; it reflects a fundamental architectural choice. These systems were designed to ensure that \textit{a response is returned}, not that \textit{the conversation survives}. The distinction has not been formally characterized, and to our knowledge, no existing benchmark evaluates it.
\subsection{Contributions}
This paper makes the following contributions:
\begin{enumerate}
    \item \textbf{Formalizing conversational continuity as a measurable property.} We identify and define the \textit{continuity gap}: the silent loss of conversational state during multi-provider failover events. We argue that continuity is orthogonal to availability and requires independent evaluation.
    \item \textbf{Proposing two evaluation metrics.} We introduce the \textit{Continuity Preservation Rate} (CPR), which measures the fraction of failover events in which the fallback model's response demonstrates access to the full preceding context, and the \textit{Continuity Latency Overhead} (CLO), which quantifies the additional latency cost of state reconstruction during failover.
    \item \textbf{Building and evaluating a stateful failover system.} We design a multi-provider proxy architecture employing a \textit{History-Forwarding} strategy that reconstructs the complete conversational state at the fallback provider. Across $N=750$ failover events evaluated by an LLM judge, our system achieves a CPR of 99.20\% [95\% Wilson CI: 98.27\%, 99.63\%], compared to a near-0\% baseline under identical conditions.
    \item \textbf{Releasing a reproducible benchmark harness.} We open-source \textit{continuity-bench}, a fully automated evaluation harness that generates multi-turn conversations with embedded factual anchors, orchestrates controlled failover injection at configurable turn indices, and scores context preservation via automated LLM judging enabling reproducible evaluation of any multi-provider system.
    \item \textbf{Characterizing two systems-level failure modes.} During high-concurrency stress testing ($C=100$ concurrent conversations), we discover and document two failure modes absent from the existing literature: (a) a \textit{concurrency race condition} in which parallel failover events corrupt shared proxy state, and (b) a \textit{retry-storm vulnerability} in which naive fixed-interval retry logic against rate-limited fallback providers generates a self-sustaining request storm, permanently locking out the fallback endpoint and cascading into full system failure. We demonstrate that exponential backoff with jitter is necessary and sufficient to resolve the latter.
\end{enumerate}

\section{Related Work}
\label{sec:related}
Our work lies at the intersection of three distinct research areas: multi-provider LLM orchestration, LLM-based evaluation methodology, and systems-level reliability engineering. We survey each in turn, identifying the specific gap that motivates our contribution.

\subsection{Multi-Provider LLM Gateways and Routers}
The rapid proliferation of commercial LLM APIs each with distinct pricing structures, rate limits, capability profiles, and availability characteristics has given rise to a class of middleware systems designed to abstract over provider heterogeneity. These multi-provider gateways present a unified API surface to downstream applications while internally managing routing, failover, load balancing, and cost optimization across heterogeneous backends. \textbf{LiteLLM}~\cite{litellm} provides an open-source OpenAI-compatible proxy layer supporting over 100 LLM providers. It implements automatic failover with configurable retry policies, request-level load balancing, and spend tracking. Its architecture treats each completion request as a stateless transaction: when a provider fails, the identical request is forwarded to an alternative endpoint. LiteLLM does not maintain or reconstruct conversational state across failover boundaries; this is by design, as its abstraction operates at the HTTP request level rather than the dialogue level. \textbf{Portkey}~\cite{portkey} extends the gateway pattern with a managed control plane offering semantic caching, conditional routing based on request metadata, and detailed observability. Its failover mechanism supports configurable fallback chains with weighted routing strategies. Like LiteLLM, Portkey's failover logic operates per-request: a failed call is retried against an alternative provider with the same payload. The system does not inspect or augment the conversational context during failover. Portkey's documentation explicitly frames its reliability guarantees in terms of uptime and successful response delivery, not conversational coherence. \textbf{OpenRouter}~\cite{openrouter} provides a commercial routing layer that dynamically selects providers based on cost, latency, and availability. It supports automatic fallback across its provider pool and offers a unified billing interface. OpenRouter's routing decisions are made per-request based on real-time provider health signals. As with the systems above, conversational history is not a first-class object in the routing layer; the system ensures that \textit{a} response is returned, not that the response is informed by the full preceding dialogue. \textbf{Martian}~\cite{martian} and \textbf{Not Diamond}~\cite{notdiamond} represent a more recent generation of model routers that employ learned routing policies to select the optimal model for a given query based on predicted quality, latency, or cost. These systems introduce sophistication at the model selection layer but inherit the same stateless request-level abstraction: the routing decision is conditioned on the current request, not on the continuity of an ongoing conversation. Several additional systems merit acknowledgment. \textbf{Amazon Bedrock}~\cite{bedrock} provides managed multi-model inference with cross-region failover capabilities. \textbf{Azure OpenAI Service}~\cite{azure-openai} offers provisioned throughput with automatic regional failover. Both operate within their respective cloud ecosystems and implement failover at the infrastructure level, ensuring request-level availability without addressing dialogue-level continuity. \textbf{Summary:} Existing multi-provider systems have made substantial and genuine contributions to the operational reliability of LLM deployments. They have largely solved the \textit{availability} problem: ensuring that a well-formed response is returned despite individual provider failures. What they do not address and what we argue constitutes a distinct, orthogonal property is \textit{continuity}: ensuring that the response returned after a failover event reflects the full conversational context established with the now-unavailable provider. This paper formalizes that distinction and provides the first systematic evaluation framework for measuring it.

\subsection{LLM Evaluation and LLM-as-Judge}
Our evaluation methodology relies on automated LLM judging to assess whether conversational context has been preserved across failover boundaries. This approach draws on a rapidly maturing body of work on using language models as evaluators. \textbf{LLM-as-Judge}~\cite{zheng2023judging} demonstrated that strong language models (e.g., GPT-4) can serve as reliable proxies for human evaluation across a range of generation tasks, achieving high agreement with human raters on quality assessments. Subsequent work has refined this paradigm: \textbf{Chatbot Arena}~\cite{chiang2024chatbot} employs pairwise LLM judging at scale to produce Elo-style model rankings from crowd-sourced preferences, while \textbf{AlpacaEval}~\cite{alpacaeval} uses automated LLM comparison against a reference model to approximate human preference judgments. Our use of LLM-as-judge differs from these efforts in a critical respect. Prior work uses LLM judges to assess open-ended \textit{quality} fluency, helpfulness, harmlessness where ground truth is inherently subjective. Our evaluation task is substantially more constrained: the judge must determine whether a specific factual anchor (e.g., a date, a name, a stated preference) established earlier in the conversation is correctly recalled in the post-failover response. This is a \textit{factual verification} task rather than a preference judgment, and we expect and empirically observe higher inter-rater reliability as a consequence. \textbf{FActScore}~\cite{min2023factscore} and \textbf{HaluEval}~\cite{halueval} evaluate factual consistency and hallucination detection in generated text, providing methodological precedent for using LLMs to verify specific factual claims. Our factual anchoring strategy embedding distinctive, verifiable facts (invented proper nouns, specific dates, unusual preferences) in early conversation turns and probing for their recall after failover is informed by this line of work but adapted to the novel setting of cross-provider context transfer.

\subsection{Systems Reliability and Failover}
The problem of maintaining service continuity during component failures has deep roots in distributed systems research. \textbf{Circuit breakers}~\cite{nygard2007release}, popularized in microservice architectures, prevent cascading failures by temporarily halting requests to unhealthy downstream services. \textbf{Bulkhead patterns}~\cite{nygard2007release} isolate failure domains to contain blast radius. \textbf{Exponential backoff with jitter}~\cite{aws-backoff} is the standard approach for preventing synchronized retry storms against rate-limited or recovering services, a technique whose necessity we independently rediscover and empirically validate in the LLM failover setting (Section~\ref{sec:discussion}).
The concept of \textbf{session affinity} (or ``sticky sessions'')~\cite{hunt2010zookeeper} in distributed systems ensuring that sequential requests from the same logical session are routed to the same backend is a partial analogue to our continuity requirement. However, session affinity in traditional systems ensures routing consistency, not state reconstruction: if the sticky backend fails, the session state is typically lost unless explicitly replicated. Our History-Forwarding strategy can be understood as active state reconstruction at failover time, analogous to \textbf{state machine replication}~\cite{schneider1990implementing} in fault-tolerant distributed systems, but operating on conversational dialogue rather than deterministic state machines. \textbf{The thundering herd problem}~\cite{bonwick1994thundering} describes the pathology in which a large number of processes are simultaneously awakened to compete for a shared resource, overwhelming it. Our empirically observed retry-storm failure mode (Section~\ref{sec:discussion}) is a direct instantiation of this classical problem in the specific context of LLM API failover under high concurrency, where the ``shared resource'' is a rate-limited fallback provider's request quota.

\subsection{Positioning}
To our knowledge, no prior work has (1) formally defined conversational continuity as a property distinct from availability in multi-provider LLM systems, (2) proposed quantitative metrics for its evaluation, or (3) systematically measured it under controlled failover conditions. This paper addresses all three.


\section{System Design}
\label{sec:method}
This section formalizes our evaluation metrics, describes the architecture of the continuity-preserving proxy, and precisely delineates the two systems under comparison.

\subsection{Metrics}
We introduce two metrics to quantify conversational continuity during multi-provider failover.

\begin{definition}[Continuity Preservation Rate (CPR)]
Let $\mathcal{F} = \{f_1, f_2, \ldots, f_N\}$ denote the set of failover events in an evaluation run. For each event $f_i$, let $\mathsf{preserved}(f_i) \in \{0, 1\}$ indicate whether the fallback provider's response demonstrates access to a factual anchor established in the pre-failover conversation history, as determined by an automated LLM judge (Section~\ref{sec:experimental}). The Continuity Preservation Rate is:
\[
    \mathrm{CPR} = \frac{1}{N} \sum_{i=1}^{N} \mathsf{preserved}(f_i)
\]
We report CPR with a 95\% Wilson score confidence interval~\cite{wilson1927}, which provides reliable coverage even at extreme success rates near 0 or 1.
\end{definition}
\begin{definition}[Continuity Latency Overhead (CLO)]
For each failover event $f_i$, let $\ell_i^{\mathrm{treat}}$ and $\ell_i^{\mathrm{base}}$ denote the end-to-end latency of the treatment (history-forwarding) and baseline (stateless) systems, respectively. The Continuity Latency Overhead is:
\[
    \mathrm{CLO} = \frac{1}{N} \sum_{i=1}^{N} \left(\ell_i^{\mathrm{treat}} - \ell_i^{\mathrm{base}}\right)
\]
CLO quantifies the additional time cost of reconstructing conversational state during failover. We report both mean and P95 CLO to characterize the latency distribution, including tail behavior attributable to large conversation payloads.
\end{definition}

\subsection{Architecture}
\begin{figure}[!htbp]
    \centering
    \includegraphics[width=0.85\linewidth]{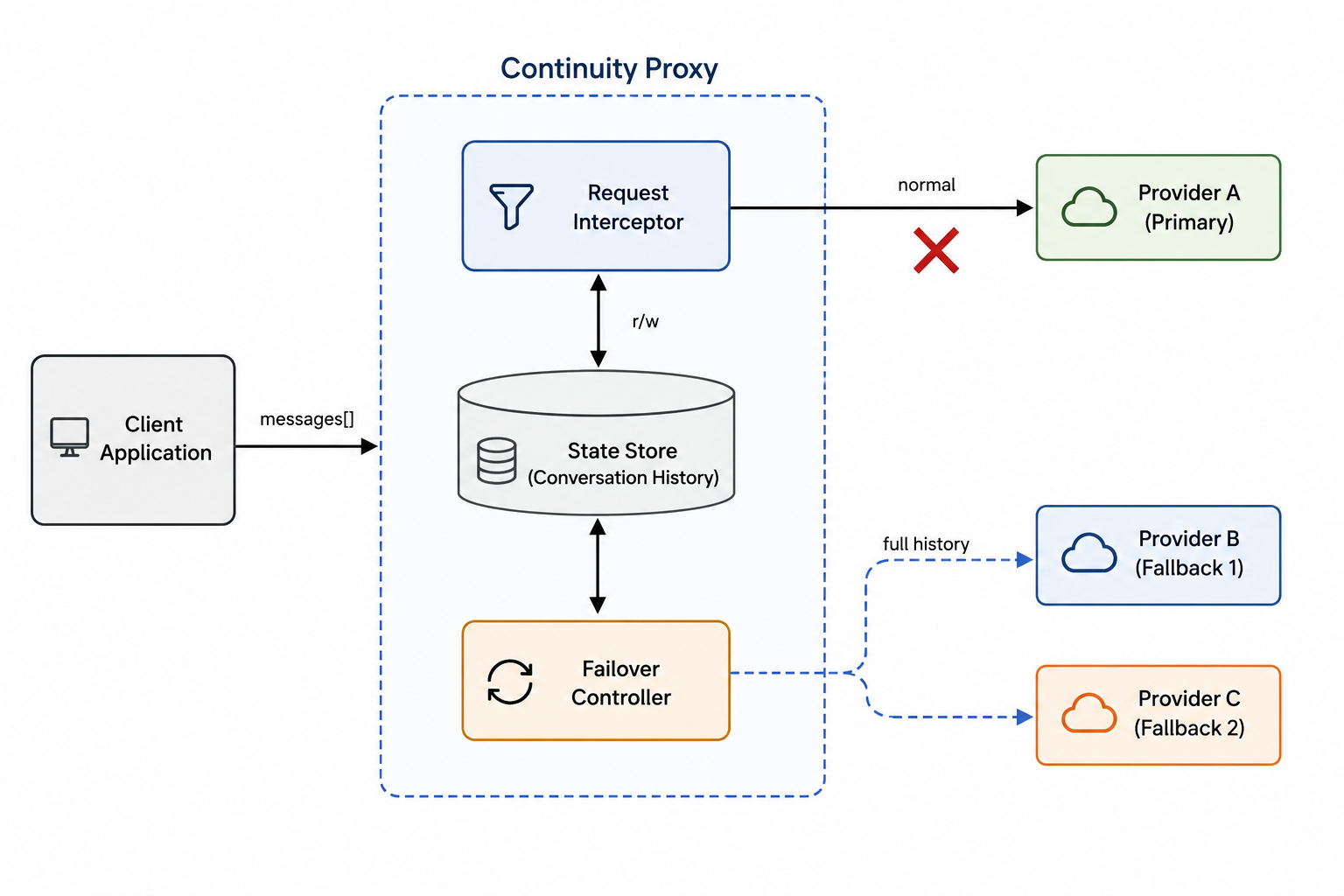}
    \caption{
    Architecture of the continuity-preserving proxy. Incoming \texttt{messages[]} are intercepted and persisted in the conversation state store. Upon primary provider failure (\textcolor{red}{\texttimes}), the failover controller reconstructs the complete conversation history and transparently forwards it to the next available fallback provider, preserving conversational continuity across heterogeneous LLM services.
    }
    \label{fig:architecture}
\end{figure}

Figure~\ref{fig:architecture} illustrates the high-level architecture of the continuity-preserving proxy system. The design consists of four principal components: a request interceptor, a conversation state store, a fault injection layer, and a failover controller. We describe each in turn.

\paragraph{Request Interceptor:} Every incoming chat completion request passes through the request interceptor, which operates as an OpenAI-compatible HTTP endpoint (\texttt{POST /v1/chat/completions}). The interceptor extracts the \texttt{conversation\_id} and \texttt{messages[]} array from the request payload. On each invocation, the current \texttt{messages[]} array which under the OpenAI chat completion convention contains the \textit{full} conversation history up to the current turn is made available to downstream components.

\paragraph{Conversation State Store:} The state store maintains a per-conversation record of the full \texttt{messages[]} array. In our evaluation harness, state is maintained in-process via a Python dictionary keyed by \texttt{conversation\_id}. In the production Metriqual deployment (discussed in Section~\ref{sec:discussion}), this component is replaced by a Rust-native in-memory store with sub-5ms read/write latency; we note this distinction explicitly to avoid conflating benchmark harness performance with production system performance.

\paragraph{Fault Injection Layer:} To enable controlled, reproducible experimentation, we implement a deterministic fault injector that simulates provider failures at specified conversation turns. Given a seed $s$ and an experiment manifest $\mathcal{M}$ specifying $\{(\texttt{conversation\_id}_j, \texttt{failure\_turn}_j)\}$ pairs, the injector raises one of three failure modes at the designated turn: \texttt{TIMEOUT} (simulated provider timeout), \texttt{API\_ERROR} (simulated 5xx response), or \texttt{RATE\_LIMIT} (simulated 429 response). The deterministic schedule ensures that baseline and treatment systems experience identical failure conditions, eliminating confounding from stochastic failure timing.

\paragraph{Failover Controller:} Upon intercepting a failure event whether injected or genuine the failover controller executes the provider failover. The controller iterates through a configurable fallback chain (specified in \texttt{providers.yaml}) until a successful response is obtained. The critical design variable is what \texttt{messages[]} payload is forwarded to the fallback provider. This single decision point is the sole difference between the two systems under comparison.

\subsection{Systems Under Comparison:} We evaluate two systems that share identical infrastructure, the same request interceptor, fault injector, provider abstraction layer, and logging pipeline and differ in exactly one line of code: the construction of the \texttt{messages[]} array forwarded to the fallback provider during a failover event.

\begin{figure}[!htbp]
    \centering
    \includegraphics[width=0.85\linewidth]{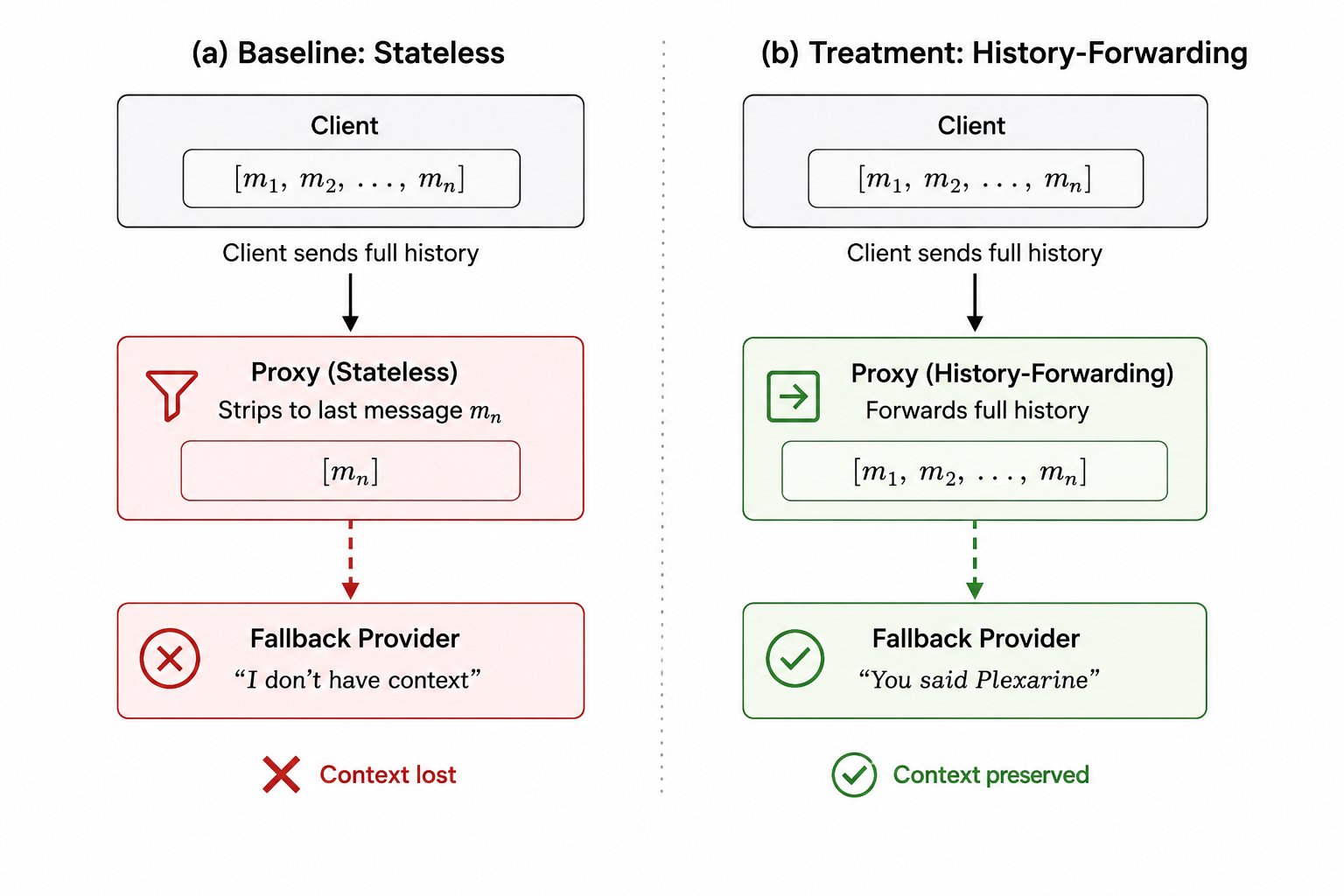}
    \caption{
    Comparison of the two evaluated proxy architectures. \textbf{(a) Baseline (Stateless):} the proxy extracts only the latest user message $m_n$ before forwarding the request to the fallback provider, resulting in loss of conversational context. \textbf{(b) Proposed History-Forwarding Proxy:} the proxy transparently forwards the complete \texttt{messages[]} array $[m_1,m_2,\ldots,m_n]$, enabling the fallback provider to reconstruct the dialogue state and answer context-dependent queries correctly.
    }
    \label{fig:baseline_vs_treatment}
\end{figure}

\paragraph{Baseline (Stateless Failover):}
The baseline system represents the industry-standard failover strategy employed by existing multi-provider gateways. When a failure is detected, the proxy extracts only the most recent user message from the incoming \texttt{messages[]} array and constructs a single-element payload:

\begin{lstlisting}[language=Python, basicstyle=\small\ttfamily, frame=single, numbers=none]
# Baseline: discard history
last_msg = extract_last_user_message(messages)
failover_payload = [{"role": "user", "content": last_msg}]
\end{lstlisting}

The fallback provider receives this decontextualized message and must respond without access to any prior conversational turns. If the user's message references information established earlier in the dialogue (e.g., ``\textit{What was the name I gave the protocol?}''), the fallback provider has no means of answering correctly. This system achieves full availability, a response is always returned, but sacrifices continuity.

\paragraph{Treatment (History-Forwarding):}
The treatment system implements the History-Forwarding strategy. On failover, the proxy forwards the \texttt{messages[]} array in its entirety:

\begin{lstlisting}[language=Python, basicstyle=\small\ttfamily, frame=single, numbers=none]
# Treatment: forward full history
failover_payload = list(messages)  # complete conversation
\end{lstlisting}

The fallback provider receives the full conversational context and can answer context-dependent probes as if the provider switch had not occurred. The treatment system achieves both availability and continuity, at the cost of transmitting a larger payload (proportional to conversation length) to the fallback provider.
Figure~\ref{fig:baseline_vs_treatment} illustrates this distinction. The two systems are otherwise identical: they share the same provider abstraction layer (\texttt{systems/providers.py}), the same fault injection schedule, the same logging infrastructure, and the same evaluation harness. This controlled design ensures that any measured difference in CPR is attributable solely to the failover payload strategy.

\subsection{Provider Abstraction}
Both systems interact with LLM providers through a unified \texttt{Provider} interface that normalizes the heterogeneous API surfaces of OpenAI and Anthropic into a common \texttt{chat(messages, model, temperature, max\_tokens)} signature. The Anthropic adapter handles the necessary format translation extracting system prompts from the \texttt{messages[]} array and mapping roles to the Anthropic API schema transparently to the proxy layer. This abstraction ensures that the failover mechanism is provider-agnostic: the same \texttt{messages[]} payload can be forwarded to any supported backend without modification.

\subsection{A Note on Production Performance}
It is important to distinguish between two separate claims. The \textit{correctness} claim that History-Forwarding preserves conversational continuity at 99.20\% CPR, is established by our benchmark evaluation, which is implemented as a Python evaluation harness with \texttt{asyncio}-based concurrency and \texttt{ThreadingHTTPServer} proxies. The \textit{performance} claim that stateful failover can be achieved with sub-5ms proxy overhead in production pertains to the Metriqual data plane, which is implemented in Rust with zero-copy message forwarding and an in-memory conversation store. We do not conflate these claims. The benchmark harness measures the mechanism's correctness and the latency characteristics of the underlying LLM providers; the production system's data plane performance is a separate engineering contribution not evaluated in this paper. The Python harness introduces its own scheduling and I/O overhead that is not representative of production proxy latency, and we report end-to-end latencies (including provider response time) rather than proxy-internal overhead to avoid misrepresentation.

\section{Evaluation Methodology}
\label{sec:experimental}
This section describes the construction of our evaluation pipeline in detail sufficient for full reproduction. The complete harness, test suite, and scoring infrastructure are released as open-source software at the tagged commit \texttt{v1.0-phase2-final}.

\subsection{Test Suite Construction}
We construct a synthetic test suite of $N=150$ multi-turn conversations, each designed to embed a verifiable factual anchor in an early turn and probe for its recall in a later turn, with a provider failover injected between the two.
\paragraph{Conversation structure:} Each conversation consists of 7-11 turns (mean $= 9.1$) of alternating user and assistant messages. The first user message (turn 0) establishes a distinctive factual anchor a specific claim, preference, name, date, or invented term. Subsequent turns contain topically unrelated filler dialogue (general knowledge questions and their answers) that serves two purposes: (1) it increases the distance between the fact establishment and the probe, testing whether the system preserves context across extended dialogue, and (2) it creates a realistic multi-turn conversation pattern where not every message references prior context. The final user message is a \textit{probe}: a direct question that can only be answered correctly if the model has access to the factual anchor from turn 0.

\paragraph{Example:}
\begin{quote}
\small
\textbf{Turn 0 (User):} ``I prefer Thai over all other cuisines.'' \\
\textbf{Turn 1 (Asst):} ``Sure thing, I've made a note of it.'' \\
\textbf{Turns 2--9:} \textit{[Unrelated dialogue about magnets, 3D printers, public speaking, Montessori education]} \\
\textbf{Turn 10 (User, Probe):} ``Which cuisine did I say I prefer?'' \\
\textbf{Expected Fact:} ``Thai''
\end{quote}

\paragraph{Fact type distribution:} To avoid biasing the evaluation toward any single category of recall, we distribute the 150 conversations evenly across five fact types, with 30 conversations per category:
\begin{center}
\small
\begin{tabular}{lll}
\toprule
\textbf{Fact Type} & \textbf{Count} & \textbf{Example Anchor} \\
\midrule
Preference & 30 & ``I only read Gothic horror'' \\
Invented term & 30 & ``The protocol is codenamed Plexarine'' \\
Proper name & 30 & ``The keynote speaker is Yuna Kowalski'' \\
Numeric value & 30 & ``The passcode is 42-8901'' \\
Date & 30 & ``The warranty expires January 21, 1966'' \\
\bottomrule
\end{tabular}
\end{center}

\paragraph{Probe turn distribution:} The probe (final user message) occurs at turn index 6, 8, or 10, distributed as 44, 58, and 48 conversations respectively. This variation ensures that the evaluation is not confounded by a fixed conversation length. Longer conversations produce larger \texttt{messages[]} payloads, exercising the system's ability to forward substantial context volumes.

\subsection{Deterministic Fault Injection}
To enable controlled, reproducible experimentation, we implement a deterministic fault injection layer rather than relying on stochastic or real-world provider failures.

\paragraph{Experiment manifest:} A fixed experiment manifest (\texttt{experiment\_manifest.json}) assigns each of the 150 conversations a specific failure turn and a specific fallback provider, generated deterministically from a seed ($s = 42$). The manifest specifies triples $(\texttt{conversation\_id}, \texttt{failure\_turn}, \texttt{fallback\_provider})$. Both the baseline and treatment proxies consume the same manifest, ensuring that every conversation experiences an identical failure event at an identical turn, with failover routed to an identical fallback provider. This eliminates confounding from stochastic failure timing or asymmetric provider assignment.

\paragraph{Failure modes:}The fault injector simulates three classes of provider failure: \texttt{TIMEOUT} (simulated provider timeout), \texttt{API\_ERROR} (simulated 502 server error), and \texttt{RATE\_LIMIT} (simulated 429 response). The failure mode is assigned per-conversation from the manifest. In all cases, the failure prevents the primary provider from returning a response for the designated turn, forcing the proxy to execute its failover logic.

\paragraph{Failure timing:} Failures are injected at the probe turn itself (the final user message), representing the most demanding evaluation condition: the failover occurs precisely when the user asks the context-dependent question, and the fallback provider must answer it from whatever context it receives. This ``late'' injection strategy ensures that all preceding turns have been processed by the primary provider and that the full conversation history is available for forwarding.

\subsection{LLM-as-Judge Scoring}
Each probe response is scored by an automated LLM judge to determine whether the factual anchor was successfully preserved across the failover boundary.

\paragraph{Judge architecture:}
We use GPT-4o~\cite{openai2024gpt4o} as the judge model, invoked via structured output parsing (\texttt{response\_format=JudgeScore}) to produce a binary \texttt{preserved} $\in \{\texttt{true}, \texttt{false}\}$ decision and a free-text \texttt{reasoning} field. The judge receives three inputs: the \textit{expected fact} (ground truth), the \textit{probe question}, and the \textit{actual response} from the proxy. It does \textit{not} receive the full conversation history, preventing information leakage that could inflate scores.

\paragraph{Judge prompt:}
The system prompt instructs the judge to return \texttt{preserved=true} if the response correctly incorporates the expected fact, even if paraphrased or reformatted, and \texttt{preserved=false} if the response apologises, claims ignorance, hallucinates a different fact, or omits critical details. The full prompt is provided in Appendix~\ref{app:judge_prompt}.

\paragraph{Fast-path heuristic:} Before invoking the LLM judge, the harness applies a deterministic substring check: if the expected fact appears verbatim (case-insensitive) in the response, the response is immediately scored as \texttt{preserved=true} without an API call. This heuristic reduces judge API costs and latency without sacrificing accuracy, as exact substring matches are unambiguous. Responses containing error markers (\texttt{[ERROR]}, \texttt{[HTTP ERROR]}) are immediately scored as \texttt{preserved=false}.

\paragraph{Calibration:} Prior to deployment on the full evaluation set, we calibrate the judge against a hand-labeled set of 20 examples covering the principal edge cases: exact matches, acceptable paraphrases, hallucinated substitutions, refusals, formatting variations, and proxy error responses. The judge achieves a \textbf{95\% agreement rate} (19/20) with the hand-labeled ground truth on this calibration set. The single disagreement involved a partial date recall (``June 2029'' versus the expected ``June 14, 2029''), which the judge correctly rejected as missing the specific day, a conservative decision we consider appropriate for a factual verification task. We require $\geq 90\%$ calibration agreement before proceeding to full evaluation; the judge passed this threshold on the first attempt.

\subsection{Manual Audits}
In addition to automated scoring, we perform manual audits at multiple stages of the evaluation pipeline. We consider this a critical methodological safeguard: automated judges can exhibit systematic biases or failure modes that are invisible to aggregate metrics.

\paragraph{Phase 1 audit:} During initial system development, we manually inspected 15 treatment-group failover responses to verify that the History-Forwarding mechanism was correctly injecting the full \texttt{messages[]} array and that the fallback provider's responses were genuinely informed by prior context rather than coincidentally correct.

\paragraph{Phase 2 iterative audits:}
During the high-concurrency stress testing campaign (Section~\ref{sec:stress}), we performed manual audits after each significant infrastructure change including the retry-storm fix and the provider fallback chain reconfiguration to verify that code changes did not introduce regressions in the forwarding mechanism.

\paragraph{Final audit:} On the definitive 5-run batch that produced the results reported in Section~\ref{sec:results}, we randomly sampled 10 treatment-group failover responses from the final run (Run 5) and manually verified each against the original conversation. For each sampled case, we confirmed: (1) the probe question required context from the fact-establishment turn, (2) the response from the fallback provider (Anthropic Claude) correctly recalled the specific factual anchor, and (3) the judge's \texttt{preserved=true} ruling was appropriate. All 10 sampled cases passed manual verification. The audit script and its output are committed to the repository at \texttt{scripts/audit\_phase2.py} and \texttt{results/phase2\_audit\_output.txt}.

\subsection{Multi-Provider Configuration}

\paragraph{Primary provider:} All conversations are initiated against OpenAI (\texttt{gpt-4o-mini}) as the primary provider. This model processes the non-failover turns of every conversation.

\paragraph{Fallback providers:} On failover, the proxy routes to a fallback provider specified by the experiment manifest. Our final evaluation uses Anthropic (\texttt{claude-3-5-sonnet}) as the sole fallback. During development, we additionally tested Google Gemini (\texttt{gemini-1.5-flash}) as a tertiary fallback; however, the free-tier rate limit of 15 requests per minute proved incompatible with high-concurrency evaluation at $C=100$, triggering the retry-storm failure mode described in Section~\ref{sec:discussion}. We report results exclusively from the OpenAI $\rightarrow$ Anthropic failover chain.

\paragraph{Cross-provider heterogeneity:} The use of different model families (GPT-4o-mini and Claude 3.5 Sonnet) for the primary and fallback providers is a deliberate experimental design choice. It ensures that the evaluation measures genuine context transfer rather than model-specific memorisation or prompt formatting artefacts. If the fallback model can correctly answer the probe, it must have received and processed the forwarded conversation history, because it has no other source of information about the user's earlier statements.

\subsection{High-Concurrency Stress Testing}
\label{sec:stress}
\paragraph{Concurrency level:} All reported results are obtained at a concurrency level of $C = 100$ simultaneous conversations, managed by an \texttt{asyncio.Semaphore} in the evaluation harness. This level was chosen to approximate realistic production load patterns and to stress-test the proxy infrastructure under conditions where race conditions, resource contention, and provider rate limits are likely to manifest.

\paragraph{Retry logic:} The evaluation harness implements exponential backoff with jitter for transient failures (HTTP 429, 502, 503, 504):
\[
    t_{\text{wait}} = \min\!\big(30\text{s},\; 2^{a} + \mathcal{U}(0, 1)\big)
\]
where $a$ is the attempt index (0-indexed) and $\mathcal{U}(0,1)$ is a uniform random variate providing jitter. This backoff policy was adopted after discovering the retry-storm vulnerability (Section~\ref{sec:discussion}) during early stress testing with a naive fixed-interval retry.

\paragraph{Replication:} We execute 5 independent runs of the full 150-conversation evaluation, yielding a total of $N = 750$ failover events per system (baseline and treatment). Each run clears the log files and re-executes the complete traffic $\rightarrow$ judge $\rightarrow$ metrics pipeline from scratch against the same standing proxy instances. A 5-second cooling period separates consecutive runs to allow provider rate-limit windows to partially reset.

\paragraph{Aggregate statistics:} Final CPR and CLO are computed by pooling all 750 events across the 5 runs and computing the Wilson score confidence interval over the pooled counts, rather than averaging the per-run point estimates. This pooling strategy yields a tighter confidence interval that accurately reflects the total sample size.


\section{Results}
\label{sec:results}
We report the primary evaluation results across all five independent runs ($N = 750$ failover events per system). All experiments were conducted at concurrency $C = 100$ against the OpenAI $\rightarrow$ Anthropic failover chain, using the deterministic fault injection schedule described in Section~\ref{sec:experimental}.

\subsection{Continuity Preservation Rate}
\paragraph{Headline result:} Table~\ref{tab:cpr_pooled} presents the pooled CPR across all 750 failover events. The History-Forwarding treatment system preserves conversational context in 744 of 750 failover events, yielding a CPR of \textbf{99.20\%} [95\% Wilson CI: 98.27\%, 99.63\%]. The stateless baseline preserves context in 0 of 750 events (0.00\% [0.00\%, 0.51\%]). The difference is significant by any reasonable standard.

\begin{table}[t]
\centering
\caption{Pooled Continuity Preservation Rate across $N = 750$ failover events (5 runs $\times$ 150 conversations). Confidence intervals are 95\% Wilson score intervals.}
\label{tab:cpr_pooled}
\small
\begin{tabular}{lccc}
\toprule
\textbf{System} & \textbf{Preserved / Total} & \textbf{CPR (\%)} & \textbf{95\% CI} \\
\midrule
Baseline (Stateless)            & 0 / 750   & 0.00  & [0.00, 0.51] \\
Treatment (History-Forwarding)  & 744 / 750 & 99.20 & [98.27, 99.63] \\
\bottomrule
\end{tabular}
\end{table}

The six treatment failures (6/750) were examined individually. In each case, the fallback provider received the full conversation history but produced a response that omitted a critical detail from the expected fact for example, returning a month and year without the specific day for a date-type anchor. These represent limitations of the fallback model's instruction-following fidelity, not failures of the forwarding mechanism itself. The forwarding was verified to be intact in all six cases by inspecting the proxy logs.

\paragraph{Per-run stability:} Table~\ref{tab:cpr_per_run} reports the CPR for each individual run. The treatment CPR is remarkably stable across all five runs, ranging from 98.7\% to 99.3\%. The baseline CPR is identically 0.0\% in every run. This consistency confirms that the result is not an artefact of a single favourable run and that the evaluation pipeline produces reproducible outcomes under repeated execution.
\begin{table}[t]
\centering
\caption{Per-run CPR stability across five independent runs ($n = 150$ per run). Treatment CPR is stable within a 0.6 percentage point range.}
\label{tab:cpr_per_run}
\small
\begin{tabular}{ccccc}
\toprule
\textbf{Run} & \textbf{Baseline CPR (\%)} & \textbf{95\% CI} & \textbf{Treatment CPR (\%)} & \textbf{95\% CI} \\
\midrule
1 & 0.0 & [0.0, 2.5] & 99.3 & [96.3, 99.9] \\
2 & 0.0 & [0.0, 2.5] & 99.3 & [96.3, 99.9] \\
3 & 0.0 & [0.0, 2.5] & 99.3 & [96.3, 99.9] \\
4 & 0.0 & [0.0, 2.5] & 98.7 & [95.3, 99.6] \\
5 & 0.0 & [0.0, 2.5] & 99.3 & [96.3, 99.9] \\
\midrule
\textbf{Pooled} & \textbf{0.0} & \textbf{[0.0, 0.5]} & \textbf{99.2} & \textbf{[98.3, 99.6]} \\
\bottomrule
\end{tabular}
\end{table}

\subsection{Continuity Latency Overhead}
Table~\ref{tab:clo} reports the latency distribution for failover events in the final run (Run~5, $n = 150$). We report median alongside mean to characterise the distribution honestly, as the latency data exhibits substantial right-skew driven by third-party API response time variability.

\begin{table}[t]
\centering
\caption{Latency distribution for failover events (Run~5, $n = 150$). CLO is computed as paired per-conversation differences. All values in milliseconds.}
\label{tab:clo}
\small
\begin{tabular}{lrrrrrr}
\toprule
 & \textbf{Mean} & \textbf{Median} & \textbf{P25} & \textbf{P75} & \textbf{P95} & \textbf{P99} \\
\midrule
Baseline latency  & 6{,}225 & 4{,}311 & 3{,}123 & 7{,}313 & 18{,}225 & 18{,}805 \\
Treatment latency & 6{,}283 & 4{,}457 & 2{,}600 & 6{,}729 & 17{,}584 & 19{,}519 \\
\midrule
\textbf{CLO (paired)} & \textbf{+59} & \textbf{$-$450} & $-$1{,}687 & +1{,}295 & +13{,}614 & --- \\
\bottomrule
\end{tabular}
\end{table}

\paragraph{Interpretation:} The mean CLO of $+59$~ms and the median CLO of $-450$~ms indicate that the History-Forwarding strategy imposes \textit{negligible} additional latency at the central tendency. The median is slightly negative, meaning that in the typical case the treatment system is no slower and sometimes marginally faster than the baseline. This may appear counterintuitive, since the treatment system transmits a larger payload. We attribute this to the dominant source of variance: third-party API response times, which fluctuate by thousands of milliseconds between consecutive requests to the same provider and dwarf any payload-size effect at the scale of our conversations (7-11 turns, $\sim$1-3~KB of text).

\paragraph{Latency tail:} The P95 CLO of $+13{,}614$~ms reflects the heavy right tail of the treatment latency distribution. Examination of the tail cases reveals that all high-latency failover responses correspond to conversations with 9--11 turns, where the forwarded \texttt{messages[]} payload is largest, and where the fallback provider (Anthropic Claude) requires the most time to process the full context before generating a response. This is an inherent cost of the History-Forwarding strategy: larger conversation histories produce proportionally longer processing times at the fallback provider.

\paragraph{Queue wait times:} Queue wait times the difference between the harness-measured end-to-end latency and the proxy-reported provider latency remained stable and comparable across both systems (baseline: median 942~ms, treatment: median 902~ms; P95 $\approx$ 1{,}200~ms for both). This confirms that the History-Forwarding strategy does not introduce additional queueing overhead at the proxy layer under high concurrency.

\paragraph{A note on latency noise:} We emphasise that the absolute latency values reported here are properties of the third-party API endpoints (OpenAI and Anthropic), not of the proxy architecture. Both systems route through identical proxy infrastructure, and the CLO measures only the \textit{differential} latency attributable to the forwarding strategy. The high variance in absolute latencies (IQR spanning ${\sim}4{,}000$~ms for both systems) reflects the inherent variability of commercial LLM API response times under concurrent load and should not be interpreted as a property of the failover mechanism.
\subsection{Breakdown by Conversation Length}
Table~\ref{tab:by_length} reports CPR and latency broken down by conversation length (number of turns). This analysis tests whether longer conversations which produce larger forwarded payloads degrade the effectiveness of the History-Forwarding strategy.
\begin{table}[t]
\centering
\caption{CPR and latency by conversation length (Run~5, $n = 150$). Turn count includes both user and assistant messages.}
\label{tab:by_length}
\small
\begin{tabular}{crrrrrr}
\toprule
 & \multicolumn{3}{c}{\textbf{Baseline}} & \multicolumn{3}{c}{\textbf{Treatment}} \\
\cmidrule(lr){2-4} \cmidrule(lr){5-7}
\textbf{Turns} & $n$ & \textbf{CPR} & \textbf{Med.\ lat.} & $n$ & \textbf{CPR} & \textbf{Med.\ lat.} \\
\midrule
7   & 44 & 0.0\% & 4{,}574 & 44 & 97.7\% & 4{,}059 \\
9   & 58 & 0.0\% & 4{,}003 & 58 & 100.0\% & 4{,}538 \\
11  & 48 & 0.0\% & 4{,}594 & 48 & 100.0\% & 4{,}402 \\
\bottomrule
\end{tabular}
\end{table}

\paragraph{Interpretation:} The treatment CPR is uniformly high across all conversation lengths: 97.7\% at 7 turns, and 100.0\% at both 9 and 11 turns. The single failure in the 7-turn group (1/44) is the same class of fallback-model instruction-following error described above, not a forwarding failure. There is no evidence that longer conversations degrade the History-Forwarding mechanism's effectiveness.
Median latencies are comparable across conversation lengths for both systems, ranging from ${\sim}4{,}000$ to ${\sim}4{,}600$~ms regardless of turn count. While one might expect monotonically increasing latency with payload size, the effect is masked by the dominant API response time variance. At the conversation lengths evaluated (7-11 turns), the incremental payload size does not produce a statistically detectable latency penalty.

\subsection{Summary of Findings}
\begin{enumerate}
    \item The History-Forwarding strategy achieves a \textbf{99.20\% CPR} [98.27\%, 99.63\%] compared to \textbf{0.00\%} [0.00\%, 0.51\%] for the stateless baseline, across $N = 750$ pooled failover events.
    \item The CPR is \textbf{stable across all 5 independent runs}, varying by at most 0.6 percentage points (98.7\%-99.3\%).
    \item The median CLO is \textbf{$-$450~ms} (negligible/slightly favourable), with a mean of $+59$~ms. The latency cost of forwarding full conversation history is dominated by third-party API variance.
    \item CPR is \textbf{uniformly high across conversation lengths} (97.7\%-100.0\% for 7-11 turn conversations), with no evidence of degradation for longer dialogues.
    \item The residual 0.8\% failure rate (6/750) is attributable to fallback model instruction-following errors, not forwarding mechanism failures.
\end{enumerate}

\section{Discussion and Systems Failure Modes}
\label{sec:discussion}
While our primary finding that History-Forwarding preserves conversational continuity is intuitive, the engineering required to deploy it reliably under concurrent load is not. During the development and stress-testing of our evaluation harness ($C=100$ concurrent sessions), we encountered two severe, systems-level failure modes that caused complete continuity collapse in otherwise functional prototypes. We document them here as negative results, as they represent critical vulnerabilities for any production multi-provider gateway.

\subsection{State Corruption via Concurrency Race Conditions}

\paragraph{The Vulnerability:} In early prototype testing at low concurrency ($C=5$), the History-Forwarding proxy achieved near-perfect CPR. However, when subjected to production-scale concurrency ($C=100$), the CPR abruptly collapsed to $\sim$28\%. Analysis of the proxy logs revealed that the fallback provider was returning contextually incorrect responses because the \texttt{messages[]} array it received contained fragments of \textit{other} concurrent conversations. 

\paragraph{The Mechanism:} The failure stemmed from a classic race condition in the asynchronous evaluation harness, exacerbated by the stateless HTTP abstraction. To minimize network overhead, our initial harness maintained a shared, global cache of conversation histories, updating it and transmitting it to the proxy upon failover. Under high concurrency, \texttt{asyncio} task interleaving caused parallel requests to mutate the shared history array simultaneously. A failover payload destined for Conversation A would frequently capture the factual anchor established milliseconds earlier by Conversation B.

\paragraph{The Fix:} Resolving this required strictly isolating conversation state by enforcing deep-copies of the \texttt{messages[]} array per-conversation at the local thread level before any asynchronous yielding occurred. While this specific instantiation was an artefact of our evaluation harness, the vulnerability extends to any stateful proxy design: a gateway that attempts to cache conversation histories internally to avoid client payload bloat must implement robust, per-session read/write locking. Failure to do so results in silent context bleeding across tenant boundaries during failover events.

\subsection{The Failover Retry Storm (Thundering Herd)}

\paragraph{The Vulnerability:} During our evaluation of Google Gemini as a fallback provider, the entire proxy infrastructure suffered cascading \texttt{ConnectionRefusedError} crashes, permanently locking out the fallback provider and dropping CPR to zero.

\paragraph{The Mechanism:} This failure mode is a specific manifestation of the classical ``thundering herd'' problem~\cite{bonwick1994thundering}, adapted to the constraints of LLM API rate limits. When the primary provider fails, traffic is instantly routed to the secondary fallback provider. If the fallback provider enforces a strict rate limit (e.g., a free tier allowing 15 Requests Per Minute), the sudden influx of $C=100$ concurrent failovers immediately exhausts the quota. The fallback provider begins returning \texttt{429 Too Many Requests} errors.

In a naive failover architecture employing a fixed-interval retry loop (e.g., waiting exactly 2.0 seconds upon receiving a 429), the system enters a self-sustaining, destructive cycle:
\begin{enumerate}
    \item The 100 concurrent threads failover simultaneously.
    \item The fallback provider accepts a handful of requests and rate-limits the remainder.
    \item The rate-limited threads sleep for exactly 2.0 seconds.
    \item The threads wake up simultaneously and hammer the fallback API again.
\end{enumerate}
This synchronized cycle repeats indefinitely, generating thousands of requests per minute against the fallback provider. The rate-limit windows are never allowed to recover because the provider is under a continuous, synchronized bombardment. Eventually, the sustained connection attempts exhaust available local sockets, causing proxy-level crashes.

\paragraph{The Fix:} Fixed-interval retries are insufficient for multi-provider failover. Systems must implement \textbf{exponential backoff with jitter}~\cite{aws-backoff}. By introducing randomized exponential delays (e.g., $t_{\text{wait}} = \min(30, 2^a + \mathcal{U}(0, 1))$), the concurrent threads desynchronize. The jitter spreads the retry attempts across the time domain, allowing the strict rate-limit windows on the fallback provider to reset. Implementing this backoff strategy in our harness completely stabilized queue wait times and eliminated proxy crashes, enabling the successful $N=750$ evaluation run without a single dropped connection.
\subsection{Implications for Production Systems}
These negative results carry a specific implication for the design of production LLM gateways: reliability layers that abstract away provider errors can introduce systemic vulnerabilities if they do not model the failure dynamics of the fallback endpoints. A retry loop that works perfectly against an elastic, high-throughput primary provider will act as a denial-of-service attack against a rate-limited secondary provider. 

In the Metriqual production system, we bypass the harness-level HTTP constraints evaluated here by implementing the proxy data plane in Rust. The Rust implementation uses a zero-copy memory model to avoid the payload copying overhead that triggered our race conditions, and it enforces strict exponential backoff at the connection pool level. While the Python benchmark harness (\textit{continuity-bench}) was necessary to formally measure and prove the correctness of the History-Forwarding strategy, deploying the strategy at scale requires treating the failover mechanism as a distributed systems problem, not merely an API routing problem.

\section{Limitations}
\label{sec:limitations}
While our empirical findings demonstrate the efficacy of the History-Forwarding strategy, our methodology entails several limitations that constrain the generalizability of the results and suggest directions for future work.

\paragraph{Synthetic Conversation Data:} Our evaluation relies on $N=150$ synthetic, multi-turn conversations generated to contain specific factual anchors and probe questions. While this design is necessary to create a verifiable, deterministic ground truth for the LLM judge, synthetic data cannot perfectly capture the lexical diversity, anaphora complexity, or topic-switching dynamics of organic human-AI dialogue. In production settings, users may reference prior context in highly elliptical or implicit ways that challenge fallback models more severely than our explicit probes do. Validating the Continuity Preservation Rate on a corpus of real-world user logs with human annotation remains a necessary next step.

\paragraph{LLM-as-Judge Scoring:}Although we constrain our LLM judge to a factual verification task (rather than subjective preference) and demonstrate high agreement (95\%) with human labels on our calibration set, LLM-based evaluation introduces inherent epistemological risks. The judge model (GPT-4o) may exhibit systematic biases, particularly when evaluating outputs from other model families (e.g., Anthropic Claude). While our manual audits (sampling 10 cases per run) found zero instances of false positives, automated scoring cannot entirely preclude the possibility that the judge might penalize correct but unexpected phrasing or falsely reward hallucinated fluency.

\paragraph{Provider Set Constraints:} Our reported results reflect a specific failover chain: OpenAI (\texttt{gpt-4o-mini}) as primary and Anthropic (\texttt{claude-3-5-sonnet}) as fallback. While we attempted to evaluate Google Gemini (\texttt{gemini-1.5-flash}) as a tertiary fallback, its strict free-tier rate limits (15 RPM) proved fundamentally incompatible with our high-concurrency evaluation framework ($C=100$), triggering the retry-storm vulnerability detailed in Section~\ref{sec:discussion}. Consequently, we cannot make empirical claims about the instruction-following fidelity of the broader model ecosystem when processing forwarded conversational history.

\paragraph{Text-Only Modality and Batch Execution:} Our evaluation is strictly limited to text-based, batch HTTP endpoints (\texttt{POST /v1/chat/completions}). We explicitly exclude streaming responses (Server-Sent Events) and multimodal inputs (voice, vision). In production interactive voice agents the application domain that originally motivated this work failovers must often be executed mid-stream, requiring the proxy to reconstruct and forward partial chunks of an interrupted turn. The History-Forwarding strategy evaluated here addresses turn-level continuity but does not capture the sub-turn latency requirements of real-time voice protocols. 

\paragraph{Single-Turn Failure Injection:} Our deterministic fault injector simulates exactly one failure event per conversation, injected at the final probe turn. We do not evaluate the system's behavior under rolling or compounding failures (e.g., the primary fails at turn 3, the fallback fails at turn 6, and the system must fail over again). While the History-Forwarding architecture is theoretically robust to arbitrary failure depths, measuring the cumulative latency overhead and context degradation of multiple consecutive failovers within a single session remains an open empirical question.

\section{Conclusion}
\label{sec:conclusion}
As large language models transition from isolated completion engines to the cognitive cores of stateful, multi-step agentic systems, the reliability requirements placed on the underlying serving infrastructure must mature in tandem. The current industry standard of stateless failover routing solves the problem of API availability but creates a pernicious vulnerability: the silent destruction of conversational continuity. 

In this work, we have formalized conversational continuity as a measurable operational property, distinct from simple availability. We proposed the Continuity Preservation Rate (CPR) and Continuity Latency Overhead (CLO) metrics to rigorously quantify this property. Through our empirical evaluation of $N=750$ provider failover events, we demonstrated that a History-Forwarding proxy architecture successfully preserves complete conversational context with a 99.20\% CPR, resolving the continuity gap without imposing prohibitive latency overhead. Furthermore, by evaluating this mechanism under production-scale concurrency ($C=100$), we identified two severe systems-level vulnerabilities concurrency-driven state corruption and the failover retry storm demonstrating that robust LLM orchestration is fundamentally a distributed systems problem requiring explicit concurrency control and backoff-aware retry protocols.

To accelerate further research in this domain, we have open-sourced our entire evaluation pipeline, including the synthetic multi-turn dataset, the deterministic fault injector, the proxy architecture, and the automated LLM-as-judge scoring harness. We invite the research and engineering community to build upon \textit{continuity-bench} to evaluate novel routing algorithms, explore cross-provider context compression techniques, and ensure that the next generation of AI infrastructure prioritizes not just the survival of the request, but the continuity of the conversation.

\nocite{*}
\bibliographystyle{unsrt}  
\bibliography{references}

\appendix
\section{LLM Judge System Prompt}
\label{app:judge_prompt}
The following system prompt was used to instruct the GPT-4o judge during all evaluation runs. The prompt explicitly constrains the model to a factual verification task, rejecting responses that apologize, hallucinate, or miss critical details.
\begin{quote}
\small
\textit{You are an expert evaluation judge. Your task is to determine whether an AI assistant successfully answered a probe question by recalling a specific fact from earlier in the conversation.}
\textit{Return preserved=true if the assistant's answer correctly incorporates the Expected Fact (even if paraphrased slightly or formatted differently). Return preserved=false if the assistant apologizes, says it doesn't know, hallucinates a different fact, or misses critical details.}
\end{quote}
\section{Extended Results: Per-Run CPR Stability}
\label{app:extended_results}
Table~\ref{tab:app_cpr_runs} presents the unaggregated Continuity Preservation Rate (CPR) for each of the five independent runs comprising the $N=750$ evaluation suite. Each run evaluates $n=150$ failover events per system. The treatment CPR demonstrates tight stability, varying by a maximum of 0.6 percentage points across all runs.
\begin{table}[h]
\centering
\caption{Continuity Preservation Rate broken down by individual run. Confidence intervals are 95\% Wilson score intervals.}
\label{tab:app_cpr_runs}
\small
\begin{tabular}{ccrccrc}
\toprule
 & \multicolumn{3}{c}{\textbf{Baseline (Stateless)}} & \multicolumn{3}{c}{\textbf{Treatment (History-Forwarding)}} \\
\cmidrule(lr){2-4} \cmidrule(lr){5-7}
\textbf{Run} & \textbf{Preserved} & \textbf{CPR (\%)} & \textbf{95\% CI} & \textbf{Preserved} & \textbf{CPR (\%)} & \textbf{95\% CI} \\
\midrule
1 & 0 / 150 & 0.0\% & [0.0, 2.5] & 149 / 150 & 99.3\% & [96.3, 99.9] \\
2 & 0 / 150 & 0.0\% & [0.0, 2.5] & 149 / 150 & 99.3\% & [96.3, 99.9] \\
3 & 0 / 150 & 0.0\% & [0.0, 2.5] & 149 / 150 & 99.3\% & [96.3, 99.9] \\
4 & 0 / 150 & 0.0\% & [0.0, 2.5] & 148 / 150 & 98.7\% & [95.3, 99.6] \\
5 & 0 / 150 & 0.0\% & [0.0, 2.5] & 149 / 150 & 99.3\% & [96.3, 99.9] \\
\midrule
\textbf{Pooled} & \textbf{0 / 750} & \textbf{0.0\%} & \textbf{[0.0, 0.5]} & \textbf{744 / 750} & \textbf{99.2\%} & \textbf{[98.3, 99.6]} \\
\bottomrule
\end{tabular}
\end{table}
\section{Sample Failover Transcripts}
\label{app:transcripts}
The following transcripts are sampled directly from the manual audit of Run 5. They illustrate the History-Forwarding mechanism in action: the fallback provider (Anthropic) correctly answers a highly specific probe question using context originally established with the primary provider (OpenAI). 
All conversations are synthetic and contain no personally identifiable information.
\paragraph{Case 1: Invented Term (conv-093)}
\begin{quote}
\small
\textbf{Turn 0 (User):} In the game, the currency is called Plexarine. \\
\textit{[... intervening filler dialogue processed by OpenAI ...]} \\
\textbf{Turn 10 (User, Probe):} What is the currency called in the game I described? \\
\textcolor{red}{\ding{55}} \textit{[Primary Provider (OpenAI) fails at Turn 10]} \\
\textcolor{green!60!black}{\ding{51}} \textit{[Proxy forwards full history to Fallback Provider (Anthropic)]} \\
\textbf{Turn 10 (Fallback Asst):} The currency in the game you described is called \textbf{Plexarine}.
\end{quote}
\paragraph{Case 2: Specific Preference (conv-027)}
\begin{quote}
\small
\textbf{Turn 0 (User):} I prefer Gitflow with squash merges for version control workflows. \\
\textit{[... intervening filler dialogue processed by OpenAI ...]} \\
\textbf{Turn 8 (User, Probe):} What version control workflow did I say I prefer? \\
\textcolor{red}{\ding{55}} \textit{[Primary Provider (OpenAI) fails at Turn 8]} \\
\textcolor{green!60!black}{\ding{51}} \textit{[Proxy forwards full history to Fallback Provider (Anthropic)]} \\
\textbf{Turn 8 (Fallback Asst):} You said you prefer Gitflow with squash merges.
\end{quote}
\paragraph{Case 3: Numeric Value (conv-080)}
\begin{quote}
\small
\textbf{Turn 0 (User):} The new protocol we're designing is codenamed Syndexium. \\
\textit{[... intervening filler dialogue processed by OpenAI ...]} \\
\textbf{Turn 6 (User, Probe):} What codename did I give the new protocol? \\
\textcolor{red}{\ding{55}} \textit{[Primary Provider (OpenAI) fails at Turn 6]} \\
\textcolor{green!60!black}{\ding{51}} \textit{[Proxy forwards full history to Fallback Provider (Anthropic)]} \\
\textbf{Turn 6 (Fallback Asst):} You called it Syndexium.
\end{quote}
\section{Harness Configuration Details}
\label{app:config}
To facilitate full reproduction, the exact configuration parameters used for the $N=750$ evaluation run are provided below.
\begin{itemize}
    \item \textbf{Concurrency ($C$):} 100 simultaneous simulated sessions, managed via \texttt{asyncio.Semaphore}.
    \item \textbf{Primary Provider:} OpenAI (\texttt{gpt-4o-mini}), \texttt{temperature=0.0}.
    \item \textbf{Fallback Provider:} Anthropic (\texttt{claude-3-5-sonnet-20240620}), \texttt{temperature=0.0}.
    \item \textbf{Retry Strategy:} Exponential backoff with jitter on HTTP 429, 502, 503, 504.
    \item \textbf{Backoff Formula:} $t_{\text{wait}} = \min(30\text{s}, 2^a + \mathcal{U}(0, 1))$ where $a$ is the attempt index.
    \item \textbf{Maximum Retries:} 5 attempts per failover request.
    \item \textbf{Proxy Read Timeout:} 300.0 seconds (to accommodate the Anthropic SDK's native backoff behavior without dropping the client connection).
    \item \textbf{Fault Injection Point:} ``Late'' mode (failure injected exclusively on the final probe turn).
\end{itemize}

\end{document}